\documentclass{article}

\PassOptionsToPackage{compress, sort, numbers, compress}{natbib}
\usepackage[preprint]{neurips_2024}

\usepackage[utf8]{inputenc} 
\usepackage[T1]{fontenc}    
\usepackage{hyperref}       
\usepackage{url}            
\usepackage{booktabs}       
\usepackage{amsfonts}       
\usepackage{nicefrac}       
\usepackage{microtype}      
\usepackage{xcolor}         
\usepackage{graphicx}
\usepackage{amsmath}
\usepackage{subcaption}
\usepackage{cleveref}
\usepackage{multicol}
\usepackage{multirow}
\usepackage{algorithm}
\usepackage{algorithmicx}
\usepackage{algpseudocode}

\definecolor{myLink}{HTML}{1F4E79}
\definecolor{myUrl}{HTML}{8E24AA}
\definecolor{myCite}{HTML}{00897B}

\hypersetup{
  colorlinks=true,   
  linkcolor=myLink,  
  urlcolor=myUrl,    
  citecolor=myCite,  
  filecolor=myLink   
}

\title{AtGCN: A Graph Convolutional Network For Ataxic Gait Detection}

\author{%
  Karan Bania \\
  Department of CS \& IS\\
  BITS Pilani, Goa Campus\\
  Goa, India - 403726
  \And
  Tanmay Tulsidas Verlekar\thanks{Corresponding author: \texttt{tanmayv@goa.bits-pilani.ac.in}} \\
  Department of CS \& IS\\
  BITS Pilani, Goa Campus\\
  Goa, India - 403726
}

\begin{document}

\maketitle

\begin{abstract}
    Video-based gait analysis can be defined as the task of diagnosing pathologies, such as ataxia, using videos of patients walking in front of a camera. 
    This paper presents a graph convolution network called AtGCN for detecting ataxic gait and identifying its severity using 2D videos. 
    The problem is especially challenging as the deviation of an ataxic gait from a healthy gait is very subtle. 
    The datasets for ataxic gait detection are also quite small, with the largest dataset having only 149 videos. 
    The paper addresses the first problem using special spatiotemporal graph convolution that successfully captures important gait-related features. 
    To handle the small dataset size, a deep spatiotemporal graph convolution network pre-trained on an action recognition dataset is systematically truncated and then fine-tuned on the ataxia dataset to obtain the AtGCN model. 
    The paper also presents an augmentation strategy that segments a video sequence into multiple gait cycles. The proposed AtGCN model then operates on a graph of body part locations belonging to a single gait cycle. 
    The evaluation results on two publicly available datasets support the strength of the proposed AtGCN model, as it outperforms the state-of-the-art in ataxia detection with an accuracy of 93.46\% and 99.5\% and in severity prediction with an MAE of 0.4169, while being 5.5
    times smaller than the state-of-the-art.
\end{abstract}

\section{Introduction}
\label{sec:intro}

The human gait, which is the forward movement of the human body through bipedal motion, requires complex coordination between the nervous, musculoskeletal, and cardio-respiratory systems~\citep{kerrigan2000refined}. 
Thus, a pathology affecting the functioning of any of these systems manifests as a deviation from a healthy gait. 
For example, cerebellar dysfunction or impaired vestibular system can cause a lack of muscle coordination and control and lead to clumsy, staggering movements with a wide-based gait called ataxic gait~\citep{Mayo-Clinic-Stuff}. 
Hence, observing the presence of ataxic gait is part of the scale for the assessment and rating of ataxia (SARA), a scale for measuring ataxia progression~\citep{SchmitzHubsch06, PerezLloret21, Hartley15, Burk09}.
Early detection of ataxia is crucial, as there is no cure for it~\citep{Mayo-Clinic-Stuff}. 
Currently, SARA is administered through visual inspection by trained professionals. 
Through the advent of artificial intelligence (AI), tasks such as detecting the presence of ataxic gait can now be performed using video recordings of a patient and machine learning (ML) algorithms~\citep{Rahman23,dataset_v2}. 
This is highly advantageous as such systems can equip general-purpose hospitals with the ability to diagnose ataxia. 
This paper improves on the state-of-the-art using Graph Convolutional Networks (GCNs) to detect ataxic gait and estimate its severity.

GCNs are a special type of Deep Learning (DL) models designed to perform inference on data described by graphs. 
They are a relatively recent development with their use-cases stretching across multiple domains, such as knowledge graph reasoning~\citep{Chung_2023, galkin2023ultra, yasunaga2022deepbidirectionallanguageknowledgegraph, zhou2024moreoneshotsubgraphreasoning} and spatiotemporal processing~\citep{sahili2023spatiotemporalgraphneuralnetworks}. 
GCNs introduce convolution to graphs of arbitrary structures, which allows them to model dynamic graphs over large-scale datasets consisting of human skeleton sequences composed of body part coordinates to perform tasks such as action recognition~\citep{yan2018spatialtemporalgraphconvolutional}. 
However, the use of GCNs for ataxic gait detection becomes challenging as:
\begin{itemize}
    \item The largest dataset for ataxic gait detection, to our best knowledge, contains only 149 video recordings~\citep{Rahman23};
    \item The variation in human skeleton sequences that allows differentiating ataxic gait from healthy gait is far more subtle than in applications such as action recognition~\citep{yan2018spatialtemporalgraphconvolutional} and fall detection~\citep{keskes21}.
\end{itemize}
The paper addresses these challenges by presenting a system that detects ataxic gait and estimates its severity using a special graph convolutional neural network called AtGCN. 
The network is an adaptation of the GCN discussed in~\citep{yan2018spatialtemporalgraphconvolutional} through fine-tuning and truncation~\citep{nevzorov22} to work on small datasets with subtle variations among its classes.
This is a significant improvement over the state-of-the-art that relies on hand-crafted features~\citep{Rahman23}.

\section{Related Work}
\label{sec:rel_work}

The literature is composed of several strategies for an objective assessment of ataxia, such as using ML algorithms to assess the speech recording of patients~\citep{kashyap20} or Magnetic Resonance Imaging (MRI) scans~\citep{Yang14}. 
The detection of ataxic gait using ML algorithms is recent and largely relies on the use of wearable sensors, such as Inertial Measurement Units (IMU), to capture gait-related features and ML algorithms, such as random forests~\citep{phan19}, and multi-layer perceptron (MLP)~\citep{lemoyne16} for classification. 
The use of depth-sensing cameras allows the capture of skeletal sequences representing the position of key body landmarks, such as hip, knee, shoulder and others, over time. 
It facilitates the extraction of features, such as trunk and head movements relative to the reference point in the pelvis and sway in the anteroposterior direction, to classify gait as either healthy or ataxic~\citep{Honda20}.

Due to privacy concerns, systems operating on 2D video camera recordings are fairly limited for ataxia detection.
Some have circumvented this issue by simulating the ataxic gait using healthy individuals. 
In~\citep{ortells18}, a dataset is created containing video recordings of 10 healthy individuals simulating ataxic gait. 
Silhouettes extracted from those video recordings are used to estimate biomechanical features such as stance phase, swing phase and step length, among others, to detect ataxic gait. 
Compact gait representations, such as gait energy image (GEI), which is generated by aggregating silhouettes belonging to a gait cycle, are used as inputs to convolutional neural networks (CNN) to detect ataxic gait in~\citep{Verlekar18}. 
While in~\cite{dataset_v2}, a sequence of key points is used to train a long short-term memory network to detect ataxic gait.
While they report very good results, the capability of the participants to simulate complex gait patterns is often questioned. 
The only public video-based dataset containing patients suffering from ataxia is presented in~\citep{Rahman23} contains only 149 video sequences. 
While the dataset is reliable, the small number of sequences prevents the use of DL models for classification. 
The work in~\citep{Rahman23} estimates biomechanical features such as step length, step width, stance phase, swing phase, speed and stability and classifies them using random forest to achieve an average accuracy of about 83\%. 
Thus, this paper explores special spatiotemporal graph convolutions and proposes a model called AtGCN that detects ataxic gait and estimates its severity. 
It also explores techniques such as truncation and fine-tuning to enable the proposed AtGCN model to be trained on such small datasets.

\section{Methodology}
\label{sec:method}
\begin{figure}[!htb]
    \centering
    \includegraphics[width=0.9\textwidth]{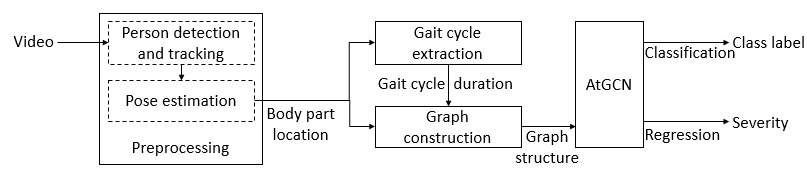}
    \caption{Proposed system architecture.}
    \label{fig:sys_arch}
\end{figure}

The system to detect ataxic gait and estimate its severity consists of 4 modules: preprocessing, gait cycle extraction, graph construction and the proposed AtGCN model, as illustrated in \cref{fig:sys_arch}. 
During preprocessing, an input video sequence is converted into a skeleton sequence composed of body part coordinates. 
The gait cycle extraction splits the skeleton sequence into multiple gait cycles. 
Next, each gait cycle is converted into a graph structure, which acts as the input for the AtGCN model. 
The proposed AtGCN model classifies the input sequence as either healthy or ataxic gait and assigns it a severity score. 

\subsection{Preprocessing}
\label{subsec:preprocess}

\begin{figure}[!htb]
    \centering
    \begin{subfigure}{0.5\textwidth}
        \centering
        \includegraphics[width=0.5\textwidth]{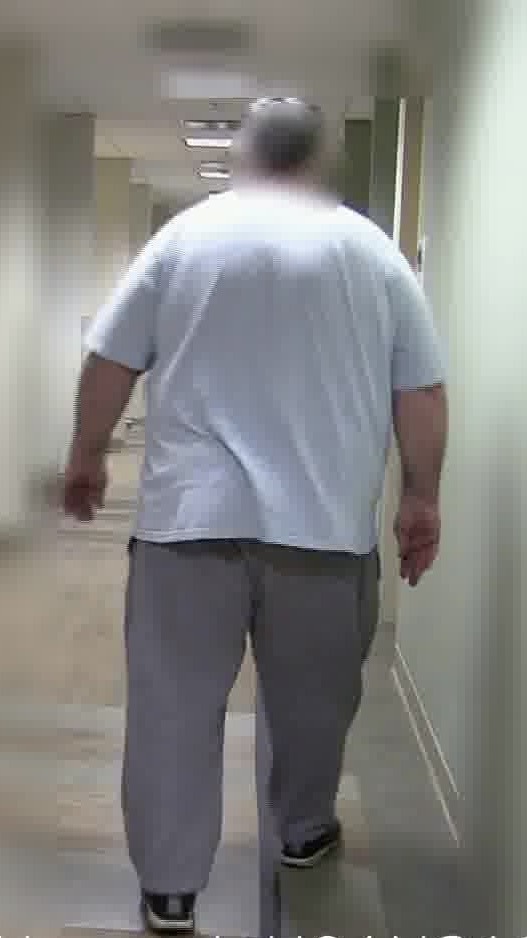}
        \caption{Input image.}
        \label{fig:plain}
    \end{subfigure}%
    \begin{subfigure}{0.5\textwidth}
        \centering
        \includegraphics[width=0.45\textwidth]{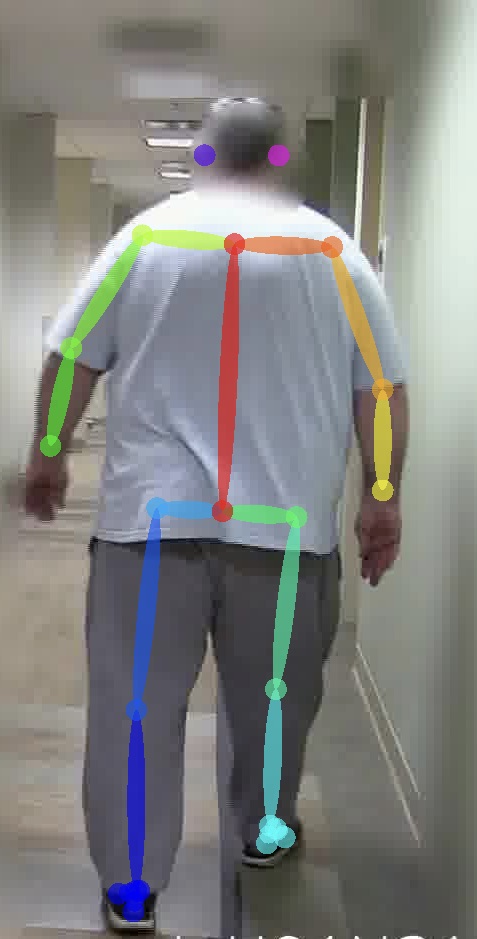}
        \caption{Confidence map and affinity field.}
        \label{fig:keypointed}
    \end{subfigure}
    \caption{2D body part coordinates estimation using OpenPose~\citep{Cao19}.}
    \label{fig:openpose}
\end{figure}

The proposed system uses a pose estimation model called OpenPose~\citep{Cao19} to estimate 18 key points over the person's body, as illustrated in \cref{fig:openpose}.
OpenPose is a NN model made up of three parts. 
The first part consists of the first 10 layers of VGG-19~\citep{simoyan15}, which captures a feature map from the input image. 
The second part consists of a multistage CNN pipeline that generates a part confidence map and part affinity field from the feature map. 
The part confidence map is the $x$, $y$ coordinate position of a particular body part in the image. 
The part affinity field encodes the location and orientation of limbs as pairwise connections between body parts. 
They are then processed using a greedy bipartite matching algorithm to estimate the pose of a person.

To track a person across frames, the proposed system uses DeepSORT~\citep{wojke17}, similar to~\citep{Rahman23}. 
It is an CNN that associates IDs to people across frames based on motion and appearance descriptors. 
The motion descriptors are estimated using Kalman filters, while the appearance descriptors are captured using a CNN.

\subsection{Gait cycle extraction}
\label{subsec:augment}
\begin{figure}
    \centering
    \includegraphics[width=0.7\textwidth]{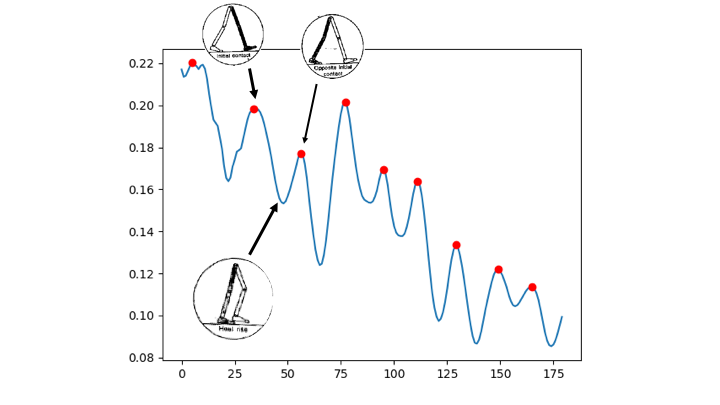}
    \caption{Plot representing the relative distance between the ankle along time.}
    \label{fig:gait_extraction}
\end{figure}

A video sequence, even six seconds long, can capture multiple gait cycles. 
A gait cycle is a functional unit of gait which is sufficient to classify it as either healthy or ataxic~\citep{Verlekar18}. 
The proposed system splits the skeleton sequence obtained using pose estimation into distinct gait cycles. 
It then uses each gait cycle as an input to the model. 
To detect a gait cycle, the proposed system analyses the distance between the feet. 
It uses the  2D coordinates of the left and right ankle to compute the Euclidean distance between the two feet. 
The distance between the feet is largest during heel strike. 
It reduces till mid-stance. 
Past mid-stance, it increases again, reaching a maximum at terminal stance. 
It then reduces till mid-swing and again reaches a maximum at the terminal swing. 
Thus, the distance, when plotted, represents a sinusoidal, where three consecutive peaks represent a gait cycle- see \cref{fig:gait_extraction}. 
To filter out noise, a Savitzky-Golay smoothing and a moving average filter are applied to the plot. 

\subsection{Graph Construction}
\label{subsec:graph_construction}

\begin{figure}[!htb]
    \centering
    \includegraphics[width=0.5\textwidth]{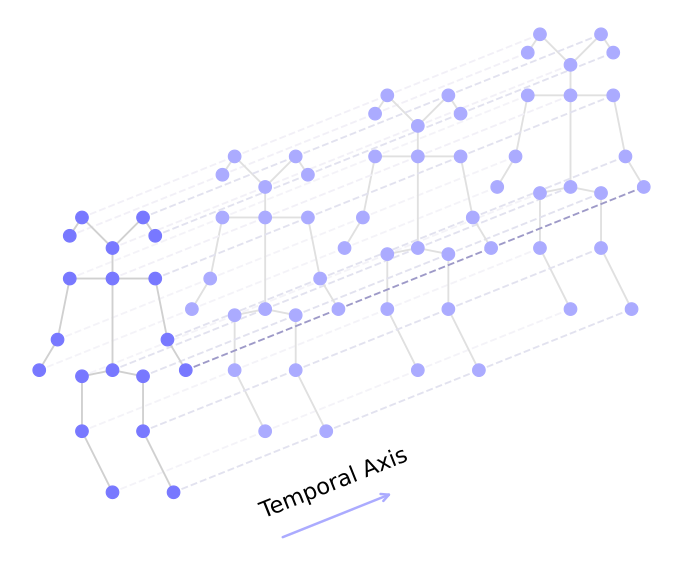}
    \caption{\label{fig:graph_construction}Input graph structure.}
\end{figure}

The 2D coordinates of the body parts captured using pose estimation over a gait cycle are turned into a spatiotemporal graph $G = (V, E)$ over a skeleton of $K$ body parts and $T$ time-steps, containing both inter- and intra-frame connections. 
The node set $V = \{v_{ti} | t = 1, ..., T, i = 1, ..., K\}$ contains all the 2D coordinates of the body parts captured over time within a gait cycle. 
The input feature vector $F(v_{ti})$ is composed of three attributes: the x and y coordinates of the body parts and the confidence of the keypoint from the OpenPose model~\citep{Cao19}.
The edges are made up of two sets called the intra-frame and inter-frame edges. 
The intra-frame edges are created according to the part affinity set $H$, such that if a connection exists between two body parts, $i$ and $j$, for a given time step $t$ then $E_{intra} = \{(v_{ti}, v_{tj}) | (i, j) \in H \}$. 
The inter-frame edges connect the same body parts across frames, i.e., $E_{inter} = \{(v_{ti}, v_{(t+1)i})\}$ - see \cref{fig:graph_construction}.

\subsection{Proposed AtGCN model}
\label{subsec:STModelling}

A GCN is a generalization over CNN, which enables them to perform convolution over non-euclidean spaces such as graphs. 
The proposed model operates on a spatiotemporal domain by applying convolution filters directly to the graph nodes and their neighbours based on a fixed sampling and weight function. 
The sampling function adopted to define the neighbour set $B(v_{ti})$ of a node $v_{ti}$ for a skeleton of body parts can be defined as: 
\begin{equation}
    \label{eq:nbrhood}
    B(v_{ti}) = \{v_{tj} | d(v_{ti}, v_{tj}) \le 1\}
\end{equation}
Where $d(v_{ti}, v_{tj})$ represents the minimum length of any path from $v_{ti}$ to $v_{tj}$. 

The work presented in~\citep{yan2018spatialtemporalgraphconvolutional} allows extending the sampling function to the spatiotemporal domain by extending the neighbour set $B(v_{ti})$ to also include temporally connected body parts. 
\begin{equation}
    \label{eq:nbrhood_temporal}
    B(v_{ti})=\{v_{qj} | (q = t, d(v_{tj}, v_{ti}) \leq D) \lor (j = i, |q - t| \leq \lfloor\Gamma/2\rfloor )\}
\end{equation}
where $\Gamma$ represents the temporal range to be included in the neighbour set. 

\begin{figure}[!htb]
    \centering
    \includegraphics[width=0.5\textwidth]{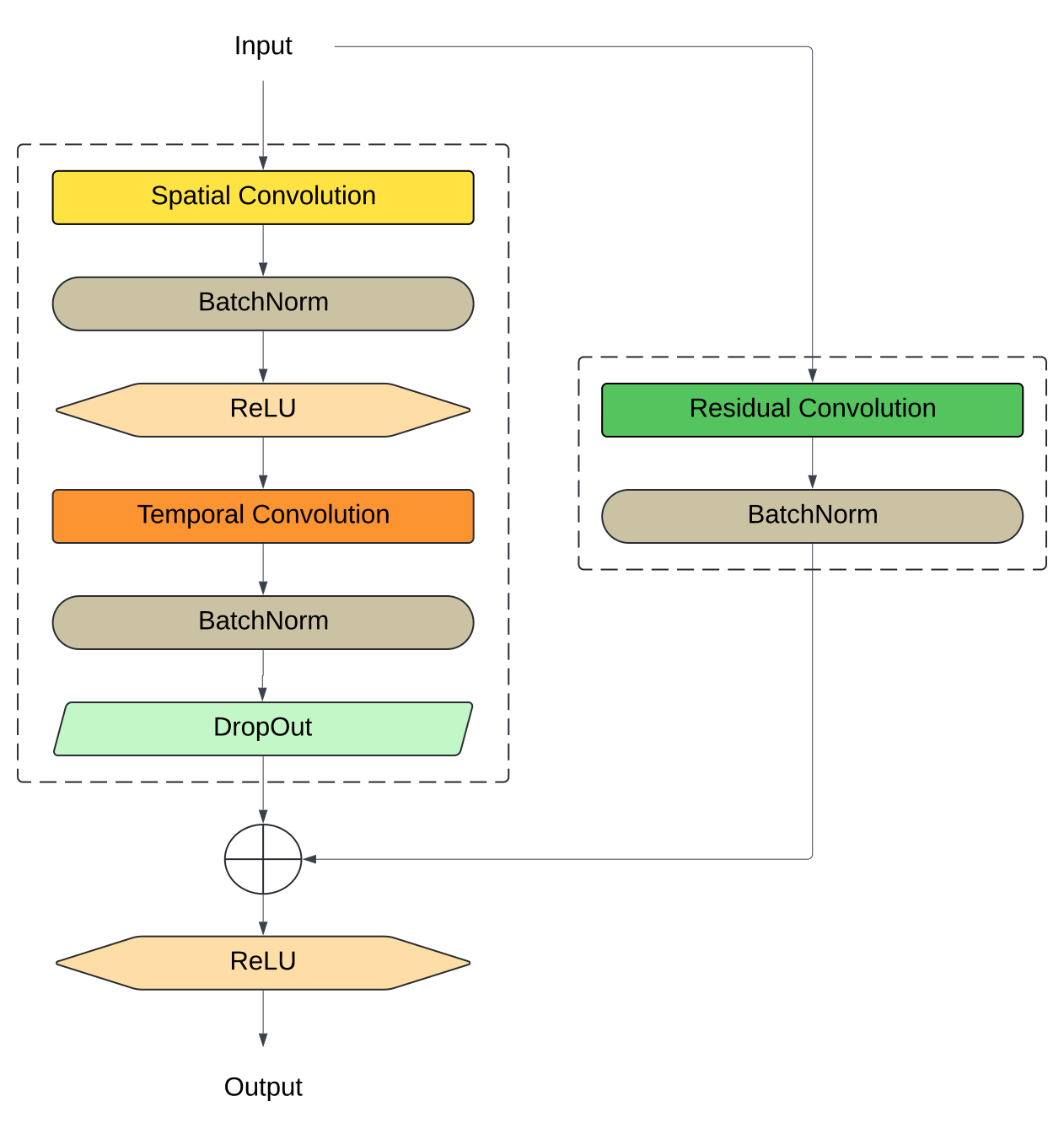}
    \caption{Spatiotemporal graph convolution block from~\citep{yan2018spatialtemporalgraphconvolutional}.}
    \label{fig:stgcn_block}
\end{figure}

The weight function requires a fixed ordering across nodes, which is achieved by partitioning the neighbour set $B(v_{ti})$ of a node $v_{ti}$ into a fixed number of $K$ labelled subsets, resulting in a mapping: 
\begin{equation}
    l_{ti} : B(v_{ti}) \rightarrow {0, ..., K-1}\label{eq:3}
\end{equation}
The partitioning strategy follows~\citep{yan2018spatialtemporalgraphconvolutional} called spatial configuration partitioning to implement the label map $l$. 
It first estimates a gravity centre, calculated as the average of all body parts coordinates in a frame. 
It then calculates the average distance $r_i$ from the gravity centre to body part $i$ across all frames in the training set. 
The mapping is then performed as follows: 
\begin{equation}
    \label{eq:label_map}
    l_{ti}(v_{tj})=
	\begin{cases}
	   0 &\mbox{if $r_{j} = r_{i}$}\\
	   1 &\mbox{if $r_{j} < r_{i}$}\\
	   2 &\mbox{if $r_{j} > r_{i}$}
	\end{cases}
\end{equation}
Since the temporal axis is well ordered, the mapping can be altered to: 
\begin{equation}
    \label{eq:label_map_temporal}
    l_{ST}(v_{qj}) = l_{ti}(v_{tj}) + (q - t + \lfloor\Gamma/2\rfloor) \times S
\end{equation}
The above equations lead to the following spatiotemporal convolution for graphs:
\begin{equation}
    \label{eq:4}
    f_{out}(v_{ti}) = \sum_{v_{tj}\in B(v_{ti})} \frac{1}{Z_{ti}(v_{tj})}f_{in}(v_{tj})\cdot \mathbf{w}(l_{ti}(v_{tj})).
\end{equation}
Where $Z_{ti}(v_{tj})$ is the normalisation term. 
The spatiotemporal convolution for the graph is implemented using a separate spatial and temporal layer within a spatiotemporal graph convolution block, as illustrated in \cref{fig:stgcn_block}.

\begin{figure}[!htb]
    \centering
    \includegraphics[width=0.8\linewidth]{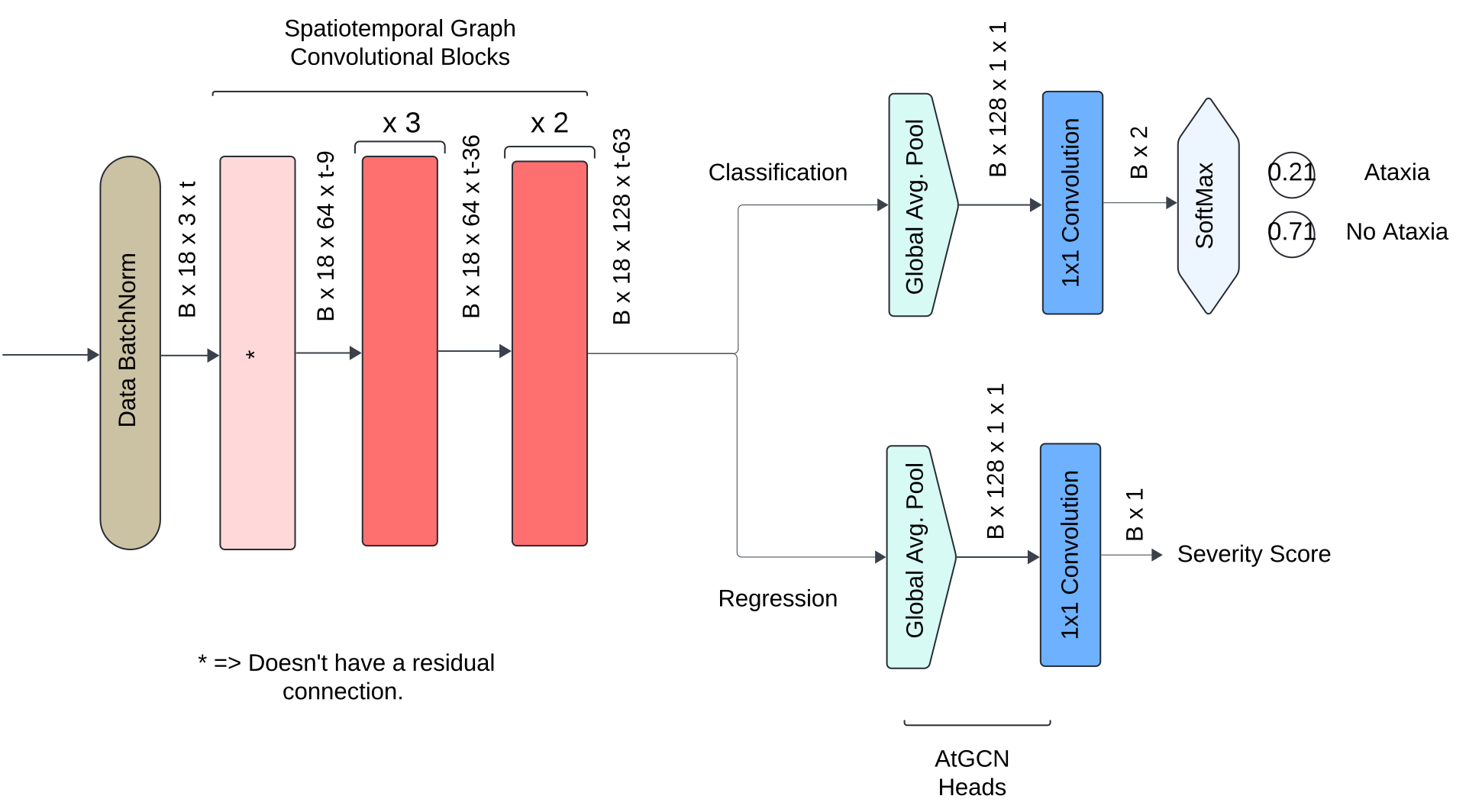}
    \caption{Proposed AtGCN model.}
    \label{fig:atgcn}
\end{figure}

The proposed AtGCN model normalises the input using batch normalisation followed by 6 spatiotemporal graph convolution blocks - see \cref{fig:atgcn}. 
The four blocks have 64 channels, followed by the next two with 128 channels with a temporal kernel size of 9. 
A dropout of 0.5 is applied after the first four blocks to avoid overfitting. 
The resulting tensor is vectorised using global average pooling followed by 1  $\times$ 1 convolution layer. The classification of gait as either ataxic or healthy is performed using the SoftMax activation. 
To estimate the severity of ataxia, global average pooling is followed by a 1  $\times$ 1 convolution layer that performs regression.
To arrive at this specific configuration of the model, an ablation study for the number of spatiotemporal graph convolution blocks is performed following \cref{alg:trunc}, whose results are reported in \cref{subsec:ablation}.

\begin{algorithm}
    \caption{Algorithm to obtain AtGCN}
    \label{alg:trunc}
    \begin{algorithmic}[1]
    \State \textbf{Input:} $M$ - pre-trained model (backbone), $Data$ - ataxia gait dataset.
    \For{$l$ = 1 to n, $n$ - total number of blocks in $M$}
    \State \textbf{Initialize:} $M$=$M$-blocks($l+1$ to $n$)
    \State $M$=Unfreeze$(M)$
    \State $M + H$, $H$- randomly initialised classification head
    \State $AtGCN(l)$=Finetune($M + H$, $Data_{train}$)
    \State Score($l$)=Predict(AtGCN($l$), $Data_{val}$)
    \EndFor
    \State $index = argmax(Score)$
    \State \textbf{Return:} $AtGCN(index)$
    \end{algorithmic}
\end{algorithm}

The pre-trained model is composed of 10 blocks of spatiotemporal graph convolution trained on the Deepmind Kinetics human action dataset~\citep{kinetics}. 
The fine-tuning is performed using stochastic gradient descent with a learning rate of $3\times10^{-05}$, a batch size of 64 and epochs set to 500. 
The proposed model is trained on a single NVIDIA V100 GPU.
An implementation can be found at \url{https://github.com/karannb/ataxia-sync}.

\section{Experiments}
\label{sec:expt}

\begin{figure}
    \centering
    \begin{subfigure}{.5\textwidth}
        \centering
        \includegraphics[width=\linewidth]{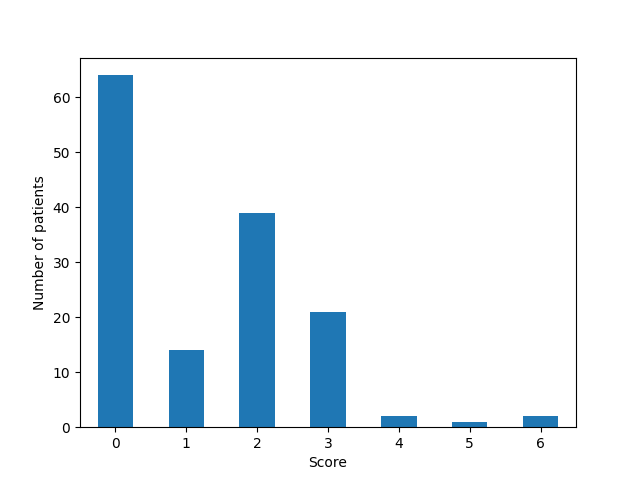}
        \caption{\centering
        Original distribution of SARA Severity score from~\citep{Rahman23}.}
        \label{fig:dist_prev}
    \end{subfigure}%
    \begin{subfigure}{.5\textwidth}
        \centering
        \includegraphics[width=\linewidth]{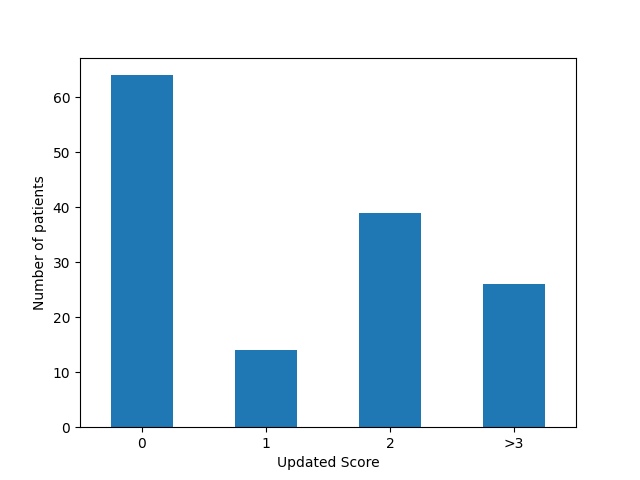}
        \caption{\centering
        Grouping of people for the severity estimation, where SARA gait score $\geq$ 3 are grouped as 3.}
        \label{fig:dist_now}
    \end{subfigure}
    \caption{Distribution of labels for the regression task.}
    \label{fig:reg_dist}
\end{figure}

The proposed AtGCN is evaluated on two datasets, the Auto-Gait dataset presented in~\cite{Rahman23} and the CA-Gait dataset introduced in~\cite{dataset_v2}.
The Auto-Gait~\cite{Rahman23} dataset is composed of 89 participants, among which 24 are healthy controls, and the remaining 65 are people with ataxia. 
The stage/degree of ataxia may vary among people with ataxia, which is indicated by its SARA gait score. 
The score ranges from 0 to 8 based on the severity of ataxia, with a score of 0 considered healthy and a score of 8 assigned to people unable to walk. 
The Auto-Gait dataset contains people with SARA scores ranging from 0 to 6, the statistics of which are presented in \cref{fig:reg_dist}. 
The participants' gait is captured multiple times to register 116 ataxic and 39 healthy videos, respectively (the public version has 6 videos less than the one in~\citep{Rahman23}). 
The dataset is captured across five sites with slight deviations in acquisition setups.
In all cases, the participants walked in a corridor with their backs facing a static camera. 
The videos capture the entire body of the participants for multiple gait cycles for exactly 6 seconds. 

The CA-Gait dataset~\cite{dataset_v2} contains recordings of 20 individuals registering both healthy and simulated ataxic gait, leading to a total of 40 videos. 
Each video recording lasts between 30 and 90 seconds, where the participants register multiple gait cycles 
as they walk towards the camera. 

The datasets are augmented by splitting each video into distinct gait cycles to further increase their size -see \cref{subsec:augment}. 
It increases the dataset size by up to threefold, allowing the training of the AtGCN model. 

To compare the results with the state-of-the-art, 20 iterations of 10-fold cross-validation, similar to~\cite{Rahman23}, are performed on the Auto-Gait~\cite{Rahman23} dataset, and special 5-fold cross-validation, similar to~\cite{dataset_v2} is performed on the CA-Gait dataset~\cite{dataset_v2}.

\subsection{Risk Prediction}
\label{subsec:risk-pred}

As reported in the \Cref{tab:risk-prediction}, the proposed AtGCN model beats the Auto-Gait \citep{Rahman23} with an improvement in accuracy and F1 score of 10\% and 13\%, respectively. 
The standard deviation (STD) of the result is significantly lower than that of the Auto-Gait \citep{Rahman23}, even when the 10-fold cross-validation is repeated over 20 iterations with different seed values.  

\begin{table*}[!ht]
    \centering
    \captionsetup{justification=centerlast}
    \setlength{\tabcolsep}{6pt}
    \renewcommand{\arraystretch}{1.25}
    \resizebox{\textwidth}{!}{
        \begin{tabular}{ c cc cc cc c }
            \toprule
            \multirow{2}{*}{Model} & \multicolumn{2}{c}{F1 score (\%) ($\uparrow$)}	& \multicolumn{2}{c}{Accuracy (\%) ($\uparrow$)} & \multicolumn{2}{c}{ROC AUC ($\uparrow$)} & Parameter \\ 
            \multicolumn{1}{c}{}	& \multicolumn{1}{c}{Mean}	& \multicolumn{1}{c}{STD} & \multicolumn{1}{c}{Mean}	& \multicolumn{1}{c}{STD} & \multicolumn{1}{c}{Mean}	& \multicolumn{1}{c}{STD} & Count ($\downarrow$) \\ 
            \midrule
            Auto-Gait~\citep{Rahman23}	& \multicolumn{1}{c}{80.23} & 9.19	& \multicolumn{1}{c}{83.06} & 6.79	& \multicolumn{1}{c}{-} & - & \multicolumn{1}{c}{-} \\ 
            GaitGraph~\citep{gaitgraph}	& \multicolumn{1}{c}{83.56} & 6.54	& \multicolumn{1}{c}{84.00} & 7.27	& \multicolumn{1}{c}{83.71} & 6.56 & \multicolumn{1}{c}{0.320M} \\ 
            AtGCN $(l=5)$	& \multicolumn{1}{c}{92.54}	& 4.39	& \multicolumn{1}{c}{92.88} & 4.41	& \multicolumn{1}{c}{92.47} & 4.50 & \multicolumn{1}{c}{0.379M} \\ 
            \textbf{Proposed AtGCN $(l=6)$} & \multicolumn{1}{c}{\textbf{93.46}} & 4.02	& \multicolumn{1}{c}{\textbf{93.76}} & 4.01 & \multicolumn{1}{c}{93.36}	& 4.19	& \multicolumn{1}{c}{0.576M} \\
            AtGCN $(l=7)$	& \multicolumn{1}{c}{92.66} & 4.54 & \multicolumn{1}{c}{92.99}	& 4.58	& \multicolumn{1}{c}{92.59} & 4.61 & \multicolumn{1}{c}{0.774M} \\
            \bottomrule
        \end{tabular}
    }
    \vspace{0.15cm}
    \caption{Comparison of the proposed AtGCN model with the state-of-the-art in classifying ataxic gait.}
    \label{tab:risk-prediction}
\end{table*}

The low performance of Auto-Gait \citep{Rahman23} can be attributed to the fact that it considers the distance between the feet and the height reduction (as the person walks away from the camera) to be the most informative features to detect ataxic gait. 
However, estimating the distance between the feet when a person is walking away from the camera can be unreliable as the video sequence lacks depth information. 
Also, the change in height features is used as an indicator of speed, but it doesn't accurately represent speed. 
The spatiotemporal graph of a gait cycle captures the entire walking pattern of a human. 
Thus, the proposed AtGCN model is far more effective in classifying between ataxic and healthy gait. 
GaitGraph~\cite{gaitgraph}, which is a lightweight GCN for gait analysis. 
While GaitGraph~\cite{gaitgraph} performs better than Auto-Gait~\cite{Rahman23}, it cannot beat the proposal AtGCN model, which achieves a 10\% improvement in the mean F1 score. 
The advantage of GaitGraph~\cite{gaitgraph} is its size of just 0.320 M parameters. 
However, as reported in \cref{tab:risk-prediction}, a comparable 5-block version of the proposed model with 0.37M parameters performs significantly better with a mean F1 score of 92\%. 
The improvement in accuracy can be attributed to the spatiotemporal graph convolution, compared to the graph convolutions in GaitGraph~\cite{gaitgraph}. 
The proposed AtGCN model captures the relationship between body parts and also captures changes in their position across time. 
In contrast, the GaitGraph~\cite{gaitgraph} aggregates the temporal component through a weighted average using $1 \times 1$ convolution. 
It should be noted that other state-of-the-art models, such as~\cite{ortells18,Verlekar18}, restrict their acquisition setup to capture gait only in the sagittal plane, where the gait dynamics are more pronounced. 
Performing ataxic gait detection in the frontal plane makes the results of the proposed AtGCN especially significant.

\begin{table*}[!ht]
    \centering
    \captionsetup{justification=centerlast}
    \setlength{\tabcolsep}{6pt}
    \renewcommand{\arraystretch}{1.25}
    \resizebox{\textwidth}{!}{
        \begin{tabular}{c cc cc c c}
            \toprule
            \multirow{2}{*}{Model} & \multicolumn{2}{c}{F1 Score (\%) ($\uparrow$)} & \multicolumn{2}{c}{Accuracy (\%) ($\uparrow$)} & \multirow{2}{*}{Fold-1 (\%) ($\uparrow$)} & Parameter \\ 
            & Mean & STD & Mean & STD & & Count ($\downarrow$) \\
            \midrule
            LSTM-based model~\cite{dataset_v2} & 97.11 & 2.70 & 96.3 & 3.46 & 99.8 & -- \\ 
            GaitGraph~\cite{gaitgraph} & 96.95 & 2.96 & 97.0 & 3.23 & -- & 0.284M \\
            AtGCN $(l=5)$ & 99.39 & 0.72 & 99.2 & 0.89 & 99.7 & 0.374M \\
            \textbf{Proposed AtGCN $(l=6)$} & \textbf{99.63} & 0.58 & \textbf{99.5} & 0.74 & 99.3 & 0.572M \\
            AtGCN $(l=7)$ & 99.46 & 0.74 & 99.3 & 0.94 & \textbf{100.0} & 0.770M \\
            \bottomrule
        \end{tabular}
    }
    \vspace{0.15cm}
    \caption{Results for the CA-Gait dataset~\cite{dataset_v2}.}
    \label{tab:ca-gait-results}
\end{table*}

To further highlight the significance of capturing the spatiotemporal information using graph structures, the proposed model is evaluated on the CA-Gait dataset~\cite{dataset_v2}. 
Fold 1 is created similar to~\cite{dataset_v2}, with each participant's ataxic and normal gait cycles distributed equally between training and validation splits. 
Other folds are generated randomly.
The proposed AtGCN beats the LSTM-based model \cite{dataset_v2}, with an improvement in accuracy of 3\%. 
It should also be noted that the proposed model has a significantly lower standard deviation across the 5-fold compared to other state-of-the-art. 
The results suggest that the AtGCN is better equipped to capture features representing Ataxic gait. 
This contrasts with the LSTM-based model \cite{dataset_v2} that effectively captures the temporal features associated with each key point but fails at capturing the spatial features between key points.

\subsection{Severity Predition}
\label{subsec:severity-pred}
A second experiment evaluates the ability of the proposed AtGCN model to estimate the severity of ataxia in the observed gait. The proposed AtGCN model performs severity prediction through regression. 
As illustrated in \cref{fig:dist_prev}, since the number of people with SARA gait scores $\geq$ 3 is quite small, the experiment considers regression across four classes: SARA score of 0, SARA score of 1, SARA score of 2 and SARA score $\geq$ 3, following ~\citep{Rahman23}.
The grouping of SARA score $\geq$ 3 is also supported by the fact that these scores necessitate medical intervention.
The second experiment is also evaluated using 10-fold cross-validation, repeated 20 times with different seed values.

\begin{table*}[!ht]
    \setlength{\tabcolsep}{6pt}
    \captionsetup{justification=centerlast}
    \renewcommand{\arraystretch}{1.25}
    \centering
    \resizebox{\textwidth}{!}{
        \begin{tabular}{ c cc cc cc c }
            \toprule
            \multirow{2}{*}{Model} & \multicolumn{2}{c }{MAE ($\downarrow$)}	& \multicolumn{2}{c }{Pearson's Coefficient ($\uparrow$)} & \multicolumn{2}{c }{MSE ($\downarrow$)}  & \multirow{2}{*}{Parameter Count ($\downarrow$)} \\ 
            \multicolumn{1}{ c }{}	& \multicolumn{1}{c }{Mean}	& \multicolumn{1}{c }{STD} & \multicolumn{1}{c }{Mean}	& \multicolumn{1}{c }{STD} & \multicolumn{1}{c }{Mean}	& \multicolumn{1}{c }{STD} & \multicolumn{1}{c }{} \\ 
            \midrule
            Auto-Gait~\citep{Rahman23}	& \multicolumn{1}{c }{0.6225} & \multicolumn{1}{c }{0.0132} & \multicolumn{1}{c }{0.7268} & \multicolumn{1}{c }{0.0144}	& \multicolumn{1}{c }{-} & -  & \multicolumn{1}{c }{-} \\
            GaitGraph~\citep{gaitgraph} & \multicolumn{1}{c }{0.8286} & \multicolumn{1}{c }{0.0998} & \multicolumn{1}{c }{0.6198} & \multicolumn{1}{c }{0.1267} & \multicolumn{1}{c }{1.0844} & \multicolumn{1}{c }{0.2413} & 0.320M \\
            AtGCN (l=5)	& \multicolumn{1}{c }{0.4249}	& 0.0689	& \multicolumn{1}{c }{0.8755} & 0.0487	& \multicolumn{1}{c }{0.3307} & 0.1164  & \multicolumn{1}{c }{0.379M} \\ 
            \textbf{Proposed AtGCN (l=6)}	& \multicolumn{1}{c }{0.4169}	& 0.0700	& \multicolumn{1}{c }{0.8738} & 0.0489	& \multicolumn{1}{c }{0.3324} & 0.1169 & \multicolumn{1}{c }{0.576M} \\ 
            AtGCN (l=7)	& \multicolumn{1}{c }{0.3878}	& 0.0776	& \multicolumn{1}{c }{0.8709} & 0.0559	& \multicolumn{1}{c }{0.3385} & 0.1360 & \multicolumn{1}{c }{0.774M} \\ 
            \bottomrule
        \end{tabular}
    }
    \vspace{0.15cm}
    \caption{Comparison of the proposed AtGCN model with the state-of-the-art in predicting the severity of ataxia.}
    \label{tab:severity-prediction}
\end{table*}

The results reported in \cref{tab:severity-prediction} suggest that the proposed AtGCN is significantly better at predicting the severity of the ataxic gait with a mean absolute error (MAE) of 0.4169 compared to the 0.6255 of the Auto-Gait \citep{Rahman23}. 
The higher MAE value of the Auto-Gait \citep{Rahman23} indicates that it finds the regression task challenging, which can be associated with the imbalance in the dataset. 
The proposed AtGCN captures the subtle variations in gait across the four classes to achieve a high Pearson's correlation with the ground truth.

\begin{figure}[h!]
    \centering
    \captionsetup{justification=centering,margin=1cm}
    \includegraphics[width=0.7\textwidth]{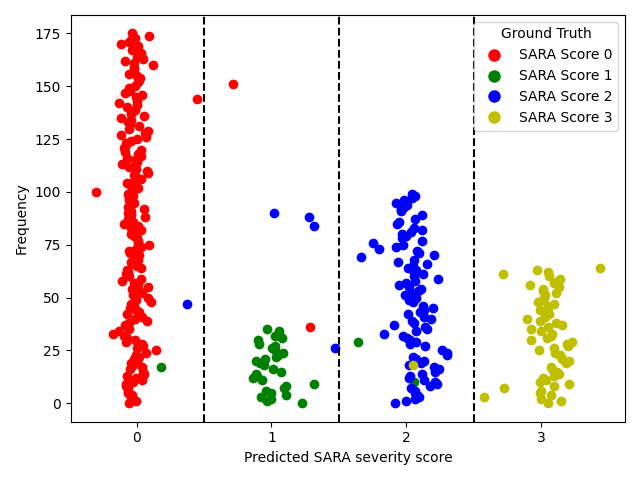}
    \caption{Binning of frequency plot used to obtain \\the confusion matrix in \cref{tab:confusion_matrix}}
    \label{fig:conf}
\end{figure}

To better understand the performance of the proposed AtGCN model against the imbalanced dataset, a confusion matrix for the regression task is reported in \cref{tab:confusion_matrix}. 
The confusion matrix is obtained by binning the severity score estimates into bins corresponding to SARA scores 0, 1, 2 and $\geq$ 3, respectively - see \cref{fig:conf}. 
The proposed AtGCN performs uniformly across all severity scores with an accuracy of more than 90\%. This can be attributed to the proposed AtGCN model's ability to capture subtle variations in gait, resulting in a high Pearson's correlation with the ground truth.

\begin{table}[h!]
\centering
\renewcommand{\arraystretch}{1.3}
\setlength{\tabcolsep}{8pt}  
    \begin{tabular}{|c|c|c|c|c|c|}
        \hline
        \multirow{2}{*}{\textbf{Ground Truth}} & \multicolumn{4}{c|}{\textbf{Prediction}} & \multirow{2}{*}{\textbf{Accuracy}} \\
        \cline{2-5}
        & \textbf{0} & \textbf{1} & \textbf{2} & \textbf{3} & \\
        \hline
        \textbf{0} & 174 & 2 & 0 & 0 & 98.86\% \\
        \cline{1-6}
        \textbf{1} & 1 & 33 & 2 & 0 & 91.67\% \\
        \cline{1-6}
        \textbf{2} & 1 & 4 & 95 & 0 & 95\% \\
        \cline{1-6}
        \textbf{3} & 0 & 0 & 1 & 64 & 98.46\%\\
        \hline
    \end{tabular}
    \vspace{0.15cm}
    \caption{Confusion Matrix for severity prediction}
    \label{tab:confusion_matrix}
\end{table}

\subsection{Ablation Study}
\label{subsec:ablation}

The final experiment is conducted to present the motivation for the proposed 6-block AtGCN model. 
The experiment is an ablation study to analyse the significance of each block following \cref{alg:trunc}. 
The results are obtained using 10-fold cross-validation for the severity prediction. 
The repetition using different seed values is skipped for this experiment.    
The results reported in \cref{tab:ablation} suggest that a few truncated blocks of spatiotemporal graph convolution from a 10-block deep network~\citep{yan2018spatialtemporalgraphconvolutional} pre-trained on Deepmind Kinetics human action dataset~\citep{kinetics} and fine-tuned on the Auto-Gait dataset~\citep{Rahman23} are sufficient to obtain results better than the state-of-the-art. 
However, the best performance is obtained using 6 blocks. 
Beyond 6, the performance of the models deteriorates. 
The change in performance can be caused by the capture of complex features associated with deeper networks. 
To capture subtle features associated with the ataxic gait, truncating the pre-trained model to 6 blocks, followed by fine-tuning, is needed. 
Thus resulting in the proposed AtGCN model.

\begin{table*}[!h]
    \setlength{\tabcolsep}{9pt}
    \renewcommand{\arraystretch}{1.25}
    \centering
    \resizebox{\textwidth}{!}{
        \begin{tabular}{ c cc cc cc c }
            \toprule
            \multirow{2}{*}{$l$} & \multicolumn{2}{c }{MAE ($\downarrow$)} & \multicolumn{2}{c }{Pearson's Coefficient ($\uparrow$)} & \multicolumn{2}{c }{MSE ($\downarrow$)} & \multirow{2}{*}{Parameter Count ($\downarrow$)} \\ 
            \multicolumn{1}{ c }{} & \multicolumn{1}{c }{Mean} & \multicolumn{1}{c }{STD} & \multicolumn{1}{c }{Mean} & \multicolumn{1}{c }{STD} & \multicolumn{1}{c }{Mean} & \multicolumn{1}{c }{STD} & \multicolumn{1}{c }{} \\ 
            \midrule
            1   & \multicolumn{1}{c }{0.8481} & 0.0956 & \multicolumn{1}{c }{0.5816} & 0.1012 & \multicolumn{1}{c }{0.9680} & 0.1924 & \multicolumn{1}{c }{0.048M} \\ 
            \midrule
            2	& \multicolumn{1}{c }{0.6286} & 0.1184 & \multicolumn{1}{c }{0.7533} & 0.1091 & \multicolumn{1}{c }{0.6306}	& 0.2337 & \multicolumn{1}{c }{0.098M} \\ 
            \midrule
            3	& \multicolumn{1}{c }{0.5191}	& 0.1114	& \multicolumn{1}{c }{0.8176} & 0.0886	& \multicolumn{1}{c }{0.4817} & 0.2143 & \multicolumn{1}{c }{0.147M} \\ 
            \midrule
            4	& \multicolumn{1}{c }{0.4416}	& 0.0707	& \multicolumn{1}{c }{0.8648} & 0.0527	& \multicolumn{1}{c }{0.3680} & 0.1390 & \multicolumn{1}{c }{0.197M} \\ 
            \midrule
            5	& \multicolumn{1}{c }{0.4151}	& 0.0623	& \multicolumn{1}{c }{0.8825} & 0.0410	& \multicolumn{1}{c }{0.3180} & 0.1069 & \multicolumn{1}{c }{0.379M} \\ 
            \midrule
            6	& \multicolumn{1}{c }{0.4093}	& 0.0682	& \multicolumn{1}{c }{\textbf{0.8838}} & 0.0395	& \multicolumn{1}{c }{\textbf{0.3114}} &  0.0937 & \multicolumn{1}{c }{0.576M} \\ 
            \midrule
            7	& \multicolumn{1}{c }{\textbf{0.3828}}	& 0.0804	& \multicolumn{1}{c }{0.8798} & 0.0351	& \multicolumn{1}{c }{0.3291} & 0.1079 & \multicolumn{1}{c }{0.774M} \\ 
            \midrule
            8	& \multicolumn{1}{c }{0.6499}	& 0.0964	& \multicolumn{1}{c }{0.7180} & 0.0716	& \multicolumn{1}{c }{0.7195} & 0.1891 & \multicolumn{1}{c }{1.497M} \\ 
            \midrule
            9	& \multicolumn{1}{c }{0.7257}	& 0.0750	& \multicolumn{1}{c }{0.6844} & 0.0740	& \multicolumn{1}{c }{0.7841} & 0.1726 & \multicolumn{1}{c }{2.286M} \\ 
            \midrule
            all blocks + FCN	& \multicolumn{1}{c }{0.7254}	& 0.0758	& \multicolumn{1}{c }{0.6844} & 0.0745	& \multicolumn{1}{c }{0.7836} & 0.1732 & \multicolumn{1}{c }{3.177M} \\ 
            \bottomrule
        \end{tabular}
    }
    \vspace{0.15cm}
    \caption{\centering Ablation study evaluating the significance of the number of spatiotemporal graph convolution blocks in the proposed AtGCN model.}
    \label{tab:ablation}
\end{table*}

\section{Limitations \& Future Work}
\label{sec:future}

This paper proposes the AtGCN model to perform the classification of gait as either healthy or ataxic and also estimate the severity of ataxia. 
The most limiting factor of this work is the availability of the data to evaluate the model. 
The Auto-Gait dataset~\citep{Rahman23}, the largest dataset to our best knowledge, has only 149 video sequences. 
The dataset size is too small to train the AtGCN model. 
A second problem with the AutoGait~\citep{Rahman23} dataset is that the 149 videos belong to 89 participants, suggesting that a few participants were recorded multiple times. 
The dataset doesn't identify these participants, thus participant-level splits were not possible to enforce, we follow the same strategy and the absolute accuracy values should be interpreted with this in mind. 
The paper addresses the first problem by presenting an augmentation technique of splitting a video into multiple gait cycles and using gait cycles to train the model. 
An experiment training the AtGCN model directly on videos for classification resulted in an accuracy of 86\%, but the loss stayed quite high. 
The second problem is addressed using a 10-fold cross-validation strategy. 
However, the dataset size still limits the capability of the model. 

Future work will consider merging datasets belonging to different gait-related pathologies to increase the size of the training set while also highlighting the capability of GCN-based models in diagnosing diverse gait-related pathologies. 
Since this is among the first applications of GCN in the clinical setting for gait analysis, there is a large scope for improvement. 
A general disadvantage of the NN models is their black-box nature of operation. 
The need for an explanation for a diagnosis is paramount in the medical field, which makes the acceptance of such models difficult. 
Thus, future work will also consider generating explanations for its decisions.

\section{Conclusion}
\label{sec:conclusion}

This paper presents a system that, given a 2D video sequence of a person walking in front of a camera, diagnoses and estimates the severity of Ataxia. 
It achieves this using the proposed AtGCN model. 
The model operates on graph structures. 
Thus, a 2D video sequence is converted into a graph structure using preprocessing, gait cycle extraction and graph construction. 
The preprocessing step converts a 2D video sequence into a sequence of 2D skeletons composed of body parts using a pose estimation model. 
The system then extracts a gait cycle using the distance between the left and right ankle. 
The spatiotemporal graph representing a gait cycle is created by connecting body parts based on anatomy and across time. 
The proposed AtGCN is composed of special spatiotemporal graph convolutions, which are defined by extending the neighbour set for the sampling and weighting functions to include temporally connected body parts.
Since it is not possible to train such a model on small datasets, the proposed AtGCN is obtained by truncating and fine-tuning a deep pre-trained spatiotemporal graph convolution model. 
Even with the truncating and fine-tuning, the proposed AtGCN model requires the augmentation of splitting a video sequence into multiple gait cycles to increase the dataset size. 
The AtGCN model beats the state-of-the-art systems, achieving 93\% and 99\% accuracies in ataxic gait detection on the two publicly available datasets and an MAE of 0.4169 in severity prediction.

\bibliography{main.bib}

\end{document}